# Brain-Inspired Spiking Neural Networks for Industrial Fault Diagnosis: A Survey, Challenges, and Opportunities

Huan Wang, Yan-Fu Li, *Senior Member, IEEE*, and Konstantinos Gryllias

*Abstract*—In recent decades, Industrial Fault Diagnosis (IFD) has emerged as a crucial discipline concerned with detecting and gathering vital information about industrial equipment's health condition, thereby facilitating the identification of failure types and severities. The pursuit of precise and effective fault recognition has garnered substantial attention, culminating in a focus on automating equipment monitoring to preclude safety accidents and reduce reliance on human labor. The advent of artificial neural networks (ANNs) has been instrumental in augmenting intelligent IFD algorithms, particularly in the context of big data. Despite these advancements, ANNs, being a simplified biomimetic neural network model, exhibit inherent limitations such as resource and data dependencies and restricted cognitive capabilities. To address these limitations, the third-generation Spiking Neural Network (SNN), founded on principles of Brain-inspired computing, has surfaced as a promising alternative. The SNN, characterized by its biological neuron dynamics and spiking information encoding, demonstrates exceptional potential in representing spatiotemporal features. Consequently, developing SNN-based IFD models has gained momentum, displaying encouraging performance. Nevertheless, this field lacks systematic surveys to illustrate the current situation, challenges, and future directions. Therefore, this paper systematically reviews the theoretical progress of SNN-based models to answer the question of what SNN is. Subsequently, it reviews and analyzes existing SNN-based IFD models to explain why SNN needs to be used and how to use it. More importantly, this paper systematically answers the challenges, solutions, and opportunities of SNN in IFD.

## I. Introduction

INDUSTRIAL fault diagnosis (IFD) is crucial for ensuring industrial equipment's stable, safe, and reliable operation. Its primary objective is to monitor the equipment's operating status and the health of each component to predict potential faults on time. Furthermore, it aims to pinpoint the fault location and identify the type of fault [1, 2]. Traditionally, this task has heavily relied on engineers with extensive knowledge and experience, capable of diagnosing engine, gearbox, and bearing faults by analyzing abnormal vibration, sound, and temperature information. However, in real-world industrial scenarios, there is a growing need for automatic fault identification and monitoring. Automated systems can significantly reduce maintenance costs, enhance the efficiency and accuracy of fault identification, and ultimately improve the overall operational safety of industrial equipment. Over the past years, the emergence of machine learning and deep learning technologies has made automated intelligent IFD a reality [3]. This transformative development has revolutionized traditional equipment maintenance practices, elevating the intelligence of industrial equipment maintenance and monitoring processes.

Intelligent IFD involves the application of machine learning and deep learning theories, such as support vector machine (SVM) [4, 5], artificial neural networks (ANN) [6], and convolutional neural networks (CNN) [7-9], for recognizing and diagnosing the health of industrial equipment. The ultimate goal is to achieve an automated fault detection and diagnosis workflow. Intelligent IFD aims to leverage advanced diagnostic models that can dynamically establish the correlation between collected data and the health status of the equipment. Creating a robust, efficient, and accurate intelligent IFD model is a significant focus of interest both in academic research and industrial applications.

In the early stages of Intelligent IFD development, machine learning-based IFD was the mainstream approach [10]. These methods typically involve two key steps: data processing and feature extraction, as well as model building and fault diagnosis. During the data processing and feature extraction step, expert knowledge is used to extract fault-related features from the collected data using signal analysis and statistical methods. The aim is to represent the original data, which is often of large capacity, with a low-dimensional feature matrix. Signal analysis methods, such as wavelet transform [11, 12] and empirical mode decomposition (EMD) [13], are commonly employed in this stage. These methods can highlight fault-related features by decomposing the signal into different components and extracting valuable information.

This work was supported by the National Natural Science Foundation of China under a key project Grant 71731008 and the Beijing Municipal Natural Science Foundation-Rail Transit Joint Research Program (L191022). (Corresponding author: Yan-Fu Li).
Huan Wang and Yan-Fu Li are with the Department of Industrial Engineering, Tsinghua University, Beijing, 100084, China (e-mail: huan-wan21@mails.tsinghua.edu.cn; liyanfu@tsinghua.edu.cn).
Konstantinos Gryllias is with the Department of Mechanical Engineering, KU Leuven, Leuven 3001, Belgium, and the Dynamics of Mechanical and Mechatronic Systems, Flanders Make, Leuven 3001, Belgium.

Next, a machine learning-based diagnostic model is constructed based on specific task characteristics to achieve intelligent industrial fault identification. Classifiers such as SVM and ANN are often the preferred choices for this step [2, 10]. These classifiers establish a mapping between the feature matrix and the fault space, enabling effective fault diagnosis. By adopting such methods, the demand for manual labor in fault diagnosis tasks is reduced, leading to the advancement of IFD into the realm of intelligent automation. However, traditional machine learning models have limited nonlinear transformation capabilities, which pose challenges in handling large-scale data volumes. Consequently, they struggle to adapt to the era of big data.

Since the 2010s, deep learning technology has made breakthroughs in various fields. With its powerful automatic feature learning and intelligent decision-making capabilities, deep learning perfectly adapts to the development of the era of big data [14, 15]. It greatly breaks through the performance bottleneck of traditional intelligent diagnosis algorithms [16-20]. In the industrial field, with the advancements in sensor technology and the Internet of Things (IoT) [21], large-scale data from industrial equipment operations is being collected. In this context, deep learning has emerged as the mainstream approach in IFD, continuously achieving breakthroughs in various application scenarios [22-26]. A plethora of deep network architectures have been proposed for different tasks. For instance, autoencoders [27] have demonstrated excellent performance in data dimensionality reduction and anomaly detection tasks. CNNs [28] have significantly reduced the parameters in traditional fully connected models, leading to excellent generalization ability. This resulted in the development of various CNN-based models, such as AlexNet [29] and ResNet [30]. To handle long-time series data, recurrent neural networks (RNNs) [31] were introduced, capable of capturing correlations between distant temporal features. Variants of RNNs, such as LSTM and GRU [32], were proposed to address the gradient vanishing and exploding issues. These advancements have greatly influenced the intelligent IFD field, leading to various results [33-35], including autoencoder-based, CNN-based, and RNN-based methods. Deep learning has significantly reduced the need for manual labor in fault diagnosis tasks, as it aims to create end-to-end intelligent decision-making models, making automated intelligent IFD a reality.

However, existing deep learning models still have challenges, including a large number of training parameters, data dependence, and high energy consumption. Moreover, their generalization and cognitive abilities are still limited. These issues remain areas of active research and development to enhance the capabilities and efficiency of deep learning models in IFD.

In recent years, the spiking neural network (SNN) [36, 37], inspired by brain science, has emerged as the third-generation neural network model. It aims to create a new generation of intelligent models with cognitive capabilities, mimicking brain neurons' basic structure and optimization methods and utilizing artificial intelligence and brain-like computing technology [38, 39]. SNN fully simulates the information transmission and dynamic characteristics between biological neurons, using spiking signals as information carriers. It aggregates valuable information flow through dynamic mechanisms between neuron synapses, allowing for feature representation of input data. As the SNN algorithm and brain-inspired computing technology continue to develop, this new computing paradigm has achieved remarkable breakthroughs. In 2019, Nature [40] reported on the hybrid Tianjic chip, developed by Tsinghua University, based on SNN and ANN models. This chip demonstrated that the cross-integration of computer science and neuroscience is the path toward artificial general intelligence. Nature [41] also highlighted the potential of SNN-based neuromorphic technology to reduce energy demand on computing platforms while achieving high intelligence. SNN has displayed excellent performance across various fields as a promising neural network model. For instance, Yin et al. [42] combined SNNs with recurrent network architectures, achieving competitive performance in speech and gesture recognition tasks. The spiking residual learning architecture proposed by Fang et al. [43] demonstrated superior performance in image recognition tasks. In the realm of fault diagnosis, scholars have begun showing interest in SNN and have developed various SNN-based diagnostic models tailored to the characteristics of IFD tasks. These research works confirmed the excellent spatial-temporal information representation potential of SNN, indicating its potential to offer a new and efficient solution for intelligent IFD.

However, there is no systematic review and comprehensive discussion in the SNN-based fault diagnosis field to illustrate the current status, challenges, and prospects. To bridge this gap, this paper aims to conduct a thorough and systematic survey of the key components of SNN-based models. It intends to provide a comprehensive overview of the current state, challenges, future developments, and opportunities associated with SNN-based diagnostic models. Specifically, the main work of this review is summarized as follows:

1) This paper comprehensively reviews IFD and SNN-based models, offering a systematic explanation of the key components and advancements in SNN-based models, thus answering what SNN is.
2) This paper systematically elucidates why SNN should be used and how to apply it effectively by combining practical IFD applications.
3) This paper outlines the roadmaps of SNN-based fault diagnosis study, offering a comprehensive understanding of the challenges, the future development direction, and potential opportunities for advancement.
4) To the best of our knowledge, this is the first systematic review paper on SNN in IFD, which is expected to provide valuable guidance for research in this field.

The rest of this paper is organized as follows. Section II focuses on the development of IFD and basic learning paradigm. Section III elaborates on key components of SNNs.

Section IV systematically answers how to implement SNNs for PHM. Section V comprehensively discusses the challenges and solutions of SNNs in PHM. Conclusions are presented in Section VI.

## II. INTELLIGENT INDUSTRIAL FAULT DIAGNOSIS

### A. Overview of Intelligent IFD Development

With the advent of the first and second industrial revolutions, human society has transitioned from traditional agricultural to modern industrial societies. This transition brought about the invention and construction of various industrial equipment, such as trains, automobiles, airplanes, and more. These technological advancements have significantly boosted productivity and led to profound societal changes. While industrial equipment greatly enhances people's quality of life, accidents involving such equipment can result in severe casualties and property losses. Hence, monitoring the health status of industrial equipment has become a matter of increasing concern for both industry and academia, evolving into an independent discipline. Fault diagnosis involves detecting and collecting information that reflects the health status of industrial equipment, while also determining fault types and degrees of failure. It aids operators in identifying abnormal equipment conditions promptly and enables targeted interventions. Traditional fault diagnosis mainly relies on threshold, physical, or signal analysis models for equipment diagnosis [44]. However, these methods heavily depend on expert experience, making it challenging to comprehensively and accurately characterize the fault characteristics of industrial equipment under dynamic and complex conditions.

Since the 1960s, the third industrial revolution has propelled human society into the digital information age, leading industrial production and equipment towards automation and informatization. Concurrently, advancements in information technology and computer technology have facilitated the deployment of various sensors on industrial equipment to monitor their operational status. This enables the collection of a substantial amount of equipment operation data, consequently driving the development of data-driven fault diagnosis technology. In practical industrial applications, different devices gather diverse sensor information, typically including speed and altitude, latitude and longitude, circuit system data (current, voltage, etc.), and mechanical transmission system information (vibration, temperature, sound, etc.). Data-driven methods [45] primarily employ effective data processing techniques to extract equipment health-related features from massive sensor data. Subsequently, classifiers and health indicators are constructed to monitor the health status of the equipment. These data-driven approaches can identify characteristic patterns of equipment failures from historical data and generalize them to current data sets, providing valuable insights for predictive maintenance and proactive fault detection.

Since the turn of the 21st century, the advancements in artificial intelligence and deep learning have marked the inception of the fourth industrial revolution, propelling human society into the era of intelligence. Concurrently, Intelligent IFD has emerged as a significant development [46]. Intelligent IFD involves the automatic and intelligent diagnosis of industrial equipment faults through artificial intelligence, deep learning, and brain-like computing technology. This approach employs intelligent models to autonomously extract fault-related knowledge from data and establish a direct correlation between the acquired knowledge and the equipment's health status. **Fig. 1** illustrates the flowchart of intelligent IFD. By delivering breakthrough performance and substantially reducing the reliance on human labor, this approach has made intelligentization the mainstream direction in fault diagnosis. Intelligent diagnosis models, such as Auto-encoders [47], CNN-based models [48, 49], and RNN-based models [50], demonstrate formidable strength and diagnostic capabilities. However, they have limitations regarding robustness, generalization, and interpretability. To address these challenges, scholars have sought inspiration from the human brain to develop novel intelligent computing models. As a result, in recent years, with the progress of brain-inspired computing, the SNN model has garnered increasing attention in fault diagnosis and showcased its distinctive advantages.

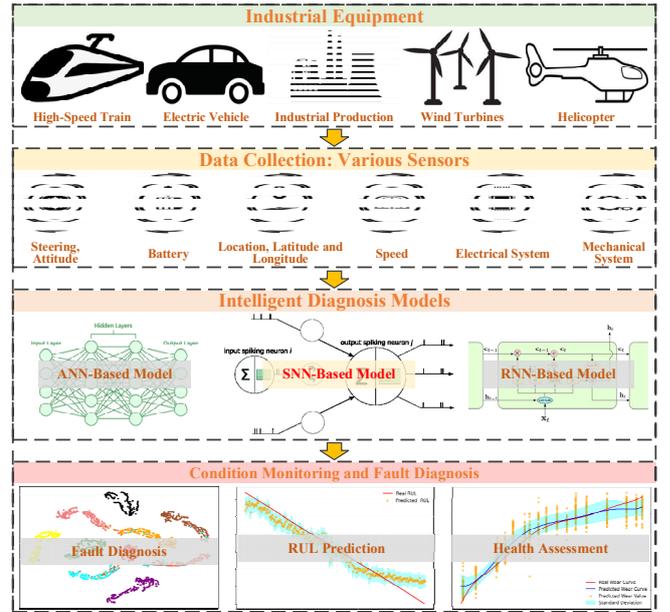

Fig. 1. The flowchart of intelligent IFD.

### B. Basic Learning Paradigm of Intelligent IFD

As depicted in **Fig. 1**, intelligent IFD is used to identify the fault type of equipment, which is typically formulated as a multi-classification task. Moreover, equipment health monitoring frequently also involves anomaly detection and regression prediction tasks. The data collected from multiple sensors is represented as $x$, and with continuous information collection over time, a training set $X = \{(x_i, y_i)\}_{i=1}^{n}$ containing a vast amount of historical data can be compiled. The learning paradigm of the intelligent diagnosis model encompasses supervised learning, semi-supervised learning, self-supervised learning, and unsupervised learning. Based on these learning paradigms, the intelligent diagnosis model can be effectively applied across various application scenarios.

Fully supervised learning refers to using manually labeled data sets to train the model to learn the mapping relationship between input and output. Given dataset $D$, each data sample $X_i$ in the dataset is accompanied by a human-annotated label $Y_i$. In the context of a training dataset $D_{train}=\{X_i,Y_i\}_{i=0}^{N}$ comprising $N$ samples, the loss is defined as follows:

$$Loss(D_{train})= \min_{\vartheta} \frac{1}{N}\sum_{i=1}^{N}\mathcal{L}(\mathcal{F}(X_i),Y_i) \quad (1)$$

$\mathcal{L}$ denotes the loss function, typically employing the cross-entropy function for fault diagnosis tasks and MSE Loss for regression prediction tasks. $\mathcal{F}$ represents the constructed PHM prediction model. While the fully supervised paradigm yields excellent performance, acquiring a sufficient number of labeled fault samples in real industrial scenarios can be costly and challenging. Consequently, learning paradigms such as semi-supervised, unsupervised, and self-supervised learning have gained wide popularity and usage to address this limitation. These paradigms leverage unlabeled data or partially labeled data to enhance model training and reduce the need for extensive labeled datasets.

Semi-supervised learning [51, 52], on the basis of supervised learning, tries to use a large number of unlabeled samples to improve the model's performance. Given a small labeled training dataset $D^l_{train}=\{X_i^l,Y_i^l\}_{i=0}^{N}$ and a large-scale unlabeled training dataset $D^u_{train}=\{X_i^u\}_{i=0}^{M}$, where $Y_i^l$ represents the label corresponding to $X_i^l$ in $D^l_{train}$, $N$ and $M$ represent the number of samples in $D^l_{train}$ and $D^u_{train}$ respectively, and the training loss is expressed as:

$$Loss(D^l_{train},D^u_{train})=$$
$$\min_{\vartheta} \frac{1}{N}\sum_{i=1}^{N}\mathcal{L}(\mathcal{F}(X_i^l),Y_i^l)+\frac{1}{M}\sum_{i=1}^{M}R(\mathcal{F}(X_i^u)) \quad (2)$$

where $R(\cdot)$ represents a task-specific function to optimize the parameter $\vartheta$ of the diagnostic model with the help of an unlabeled dataset.

Unsupervised learning [53] is dedicated to learning the statistical regularity or internal structure of data from unlabeled data, which uses unlabeled dataset $D^u_{train}=\{X_i^u\}_{i=0}^{M}$ for training and learning. The basic idea of unsupervised learning is to compress the given data to find the potential structure of the data. It assumes that the result of the compression with the smallest loss is the most essential structure. Clustering and dimensionality reduction are two basic unsupervised algorithms that learn the underlying structure of data to divide data into different classes or reduce data dimensions.

Self-supervised learning [54] aims to use the data's inherent structure and statistical laws to automatically construct pseudo-labels and pretext tasks to learn valuable features from unlabeled data. Given an unlabeled dataset $D^u_{train}=\{X_i^u\}_{i=0}^{M}$, the corresponding pseudo label $P_i$ is automatically constructed based on the characteristics of the data itself to obtain the dataset $D^p_{train}=\{X_i^p,P_i\}_{i=0}^{M}$. Based on the above dataset, the training process is expressed as follows:

$$Loss(D^p_{train})= \min_{\vartheta} \frac{1}{M}\sum_{i}^{M}\mathcal{L}(\mathcal{F}(X_i^p),P_i) \quad (3)$$

It is evident that the key advantage of self-supervised learning lies in its ability to leverage a vast amount of readily available and inexpensive unlabeled data for training and learning. By doing so, self-supervised learning can extract valuable features without relying heavily on costly labeled datasets.

### C. Fault Diagnosis Encounters Spiking Neural Networks

The SNN originated from the biological neural network with the initial goal of simulating its potential mechanisms. Over time, it has undergone gradual optimization, leading to improved performance, which has made it increasingly attractive for computing purposes. Bionic computing models typically utilize bionic neuron models to construct neural network topologies for specific tasks. Compared to artificial neurons, the neuron models adopted by SNN exhibit more realistic bionic characteristics and complex dynamic mechanisms. As a result, these neuron models may endow SNN with more robust learning abilities and problem-solving capabilities.

As discussed in Section II.A, the advent of deep learning technology has propelled the field of fault diagnosis into the era of intelligence. Nevertheless, the current level of intelligence provided by deep learning remains limited, significantly lower than that of the biological brain. Therefore, the future focus lies in further enhancing the model's intelligence to match or surpass the biological brain's performance. SNN offers a promising solution by mimicking the biological brain and harnessing its advantages to approach its performance level. While the biological brain is an extraordinarily complex and intricate system, the existing SNN's performance is still far from matching it. Nonetheless, this promising direction will undoubtedly continue to drive the development of intelligent computing and fault diagnosis. Hence, when fault diagnosis encounters SNN, it presents a promising and efficient solution for practical applications in fault diagnosis.

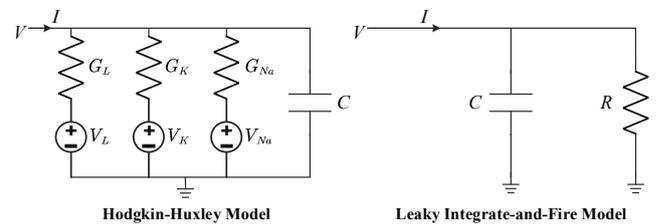

Fig. 2. The equivalent circuit diagram of the Hodgkin-Huxley model and Leaky Integrate-and-Fire Model.

### III. KEY COMPONENTS OF SPIKING NEURAL NETWORKS

#### A. Bionic Spiking Neuron Model

*1) Hodgkin-Huxley Model*

The Hodgkin-Huxley model [55] is a widely-used mathematical model that effectively describes the behavior of biological neurons. It was proposed by Hodgkin and Huxley, who were awarded the Nobel Prize in Physiology for their groundbreaking work. **Fig. 2** presents the equivalent circuit diagram of the Hodgkin-Huxley model, and its dynamic expression can be represented as follows:

$$I = C\frac{dV}{dt} + G_{Na}m^3h(V-V_{Na}) + G_K n^4(V-V_K) + G_L(V-V_L) \quad (4)$$

where $G_{Na}$, $G_K$, and $G_L$ are the model conductance parameters for sodium, potassium, and leakage channels, respectively. $V_{Na}$, $V_K$, and $V_L$ are the reverse potentials for these channels. *m*, *n*, and *h* are three gating variables, which can be described by the following formulas:

$$\frac{dm}{dt} = \alpha_m(V)(1-m) - \beta_m(V)m \quad (5)$$

$$\frac{dn}{dt} = \alpha_n(V)(1-n) - \beta_n(V)n \quad (6)$$

$$\frac{dh}{dt} = \alpha_h(V)(1-h) - \beta_h(V)h \quad (7)$$

where $\alpha_m(V)$, $\alpha_n(V)$, $\alpha_h(V)$, $\beta_m(V)$, $\beta_n(V)$, and $\beta_h(V)$ are functions that are empirically determined, and they represent the rate of entry and exit of sodium and potassium ions into and out of the cell membrane. The Hodgkin-Huxley model meticulously models the behavior of sodium and potassium ion channels and accurately describes the changes in action potentials of neurons. It provides a detailed characterization of the dynamic mechanisms involved in these processes.

*2) Leaky Integrate-and-Fire Model*

Although the Hodgkin-Huxley model can accurately describe the dynamic changes of neurons, its computational complexity is high, and it is challenging to build a large-scale neural network model. Therefore, the Leaky Integrate-and-Fire (LIF) model [56] was proposed, which is a simplified version of the Hodgkin-Huxley model and retains the essential characteristics of neuron dynamics. It can be described as:

$$\tau_m \frac{dV}{dt} = -(V - V_{rest}) + R_m I(t)$$
$$\text{if } V \geqslant \vartheta \text{ then } \lim_{\alpha \to 0; \alpha > 0} V(t+\alpha) = V_r \quad (8)$$

where $V$ and $V_{rest}$ represent the neuron's membrane potential and resting potential, respectively. $\tau_m$ represents the membrane time constant, $I(t)$ represents the input current, and $R_m$ represents the leak resistance of the cell membrane. $\vartheta$ represents the discharge threshold, and $V_r$ represents the reset potential. The LIF model significantly simplifies the complexity of the neuron model while preserving its three key features: membrane potential accumulation, leakage, and threshold discharge process. As a result, the LIF model is commonly used to construct deep SNN models that are more amenable to optimization and training, while still being biologically plausible.

*3) Izhikevich Model*

The Hodgkin-Huxley model [57] has high biological rationality but complex calculations, and the LIF model has high computational efficiency but low biological rationality. The Izhikevich model hopes to combine the advantages of these two models as much as possible to achieve a model with high biological rationality and computational efficiency. It can be described as:

$$\frac{dV}{dt} = 0.04V^2 + 5V + 140 - U + I$$
$$\frac{dU}{dt} = a(bV - U) \quad (9)$$

where *U* represents the membrane potential recovery variable, which effectively describes the behavior of ionic current in general. Parameters *a* and *b* are utilized to adjust the time scale of *U* and its sensitivity concerning the membrane potential *V*, respectively. The model can accurately simulate the firing patterns of a wide range of neurons in the brain, while the computational overhead required for its simulation is only about one-hundredth of that required by the Hodgkin-Huxley model.

*B. Information Encoding of SNN-based Models*

The encoding of information in SNN-based models is a crucial research area, and two widely used methods are rate encoding and temporal encoding. Rate encoding represents information through the spiking firing rate, which indicates the average number of spikes fired by a neuron during a specific recording time. Stronger stimuli generally lead to higher spike firing rates, indicating the level of neuron excitement. Rate encoding is extensively used in various SNN models due to its simplicity in encoding and decoding. However, rate encoding's simplicity might not fully capture the complex information processing in the brain. To address this limitation, the temporal encoding mechanism [58, 59] improves rate encoding by considering not only the firing frequency of spikes but also the interspike interval between two spikes. Temporal encoding focuses on the time pattern of spike emissions, providing greater flexibility in encoding complex feature information. As a result, it exhibits distinct advantages over rate encoding in terms of information transmission efficiency. Moreover, based on research on brain neurons, additional encoding methods such as bursting encoding and population encoding have been proposed, offering effective encoding approaches for neuron information transmission. These diverse encoding techniques provide valuable options for SNN-based models to represent and process information effectively.

*C. Network Topology of SNN-based Models*

Once spike neuron modeling and information encoding are achieved, neurons need to be interconnected in specific topologies to collaborate in building complex neural network architectures to accomplish specific tasks. Similar to the ANN model, SNN also encompasses a variety of topological structures. For instance, in the spiking FCNs [60], neurons are densely connected, and each neuron receives spike outputs from all neurons in the previous layer. Spiking RNNs [61] incorporate feedback connections, facilitating the formation of recurrent paths for spiking signals in the network. This makes them more suitable for processing time-dependent tasks, such as sequence processing and dynamic pattern recognition. The spiking CNN [62, 63] introduces the convolutional encoding mechanism, enabling local convolution operations between learnable filters and input pulses. This architecture is adept at handling data with spatial structures. In spiking convolution, the weights of the filters model synaptic connections' strengths, and the convolution operation simulates synaptic transmission and activation of postsynaptic neurons. Beyond these, complex SNN-based topological structures have been proposed, such as the SNN model based on the attention mechanism and the SNN model based on the Transformer architecture. These advances demonstrate that the SNN-based model, similar to the ANN-based model, can achieve various complex network architectures with different capabilities through distinct neuron connection methods. This enhances the flexibility and application scope of the SNN model.

### D. Optimization Algorithms of SNN-based Models

#### 1) Spike Time-Dependent Plasticity (STDP)

Research has demonstrated that alterations in the synaptic strength of neurons are contingent upon the precise timing of presynaptic and postsynaptic spikes. When the postsynaptic spike transpires shortly after the presynaptic spike, it reinforces synaptic connections, giving rise to what is known as long-term potentiation (LTP). Conversely, if the postsynaptic spike precedes the presynaptic spike, the synapse undergoes a process called long-term depression (LTDP). STDP [64, 65] encapsulates this mechanism, wherein the temporal alignment of presynaptic and postsynaptic events precipitates modifications in the efficacy of synaptic connections between pre- and postsynaptic neurons. This phenomenon can be formulated as follows:

$$\triangle \omega = \sum_n \sum_m K(t_{post}^m - t_{pre}^n) \quad (10)$$

where $t_{post}^m$ and $t_{pre}^n$ denote postsynaptic and presynaptic spike times, respectively. The kernel function $K(x)$ is expressed as:

$$K(x) = \begin{cases} A^+ \exp(-x/\tau^+), & x > 0 \\ A^- \exp(x/\tau^-), & x < 0 \end{cases} \quad (11)$$

where $\tau^+$ and $\tau^-$ are respectively used to control the decay of the exponential function; $A^+$ and $A^-$ indicate the direction of weight change of the corresponding synapse.

#### 2) Remote Supervision Method (ReSuMe)

ReSuMe [66] is an efficient learning rule for SNNs that aims to force SNNs to output desired spike trains. Its learning rules can be expressed as follows:

$$\frac{d\omega_{oi}}{dt} = (S_d(t) - S_o(t))\left(a_d + \int_0^\infty a_{di}(s) S_i(t-s) ds\right) \quad (12)$$

where $\omega_{oi}$ denotes the synaptic weights and $S_i(t)$, $S_o(t)$ and $S_d(t)$ are the input, output and desired spike train. $a_d$ is an uncorrelated factor; $a_{di}(s)$ is a kernel function used to specify the learning window.

#### 3) SpikeProp

The SpikeProp algorithm [67] employs error backpropagation to train an SNN-based model to learn target spikes, which expect to minimize the difference between actual and expected spike times. It can be expressed as:

$$E = \frac{1}{2} \sum_j (t_j^a - t_j^d)^2 \quad (13)$$

where $t_j^a$ and $t_j^d$ represent the actual and desired times of spikes firing, respectively. Analogous to the gradient descent technique employed in ANN models, the adjustment of synaptic weights adheres to the subsequent formula:

$$\triangle \omega_{ij}^k = -\alpha \frac{\partial E}{\partial \omega_{ij}^k} \quad (14)$$

where $\alpha$ represents the learning rate, and $\partial E / \partial \omega_{ij}^k$ can be derived using the chain rule.

#### 4) ANN-to-SNN

The ANN-to-SNN paradigm [68, 69] represents a strategic approach that capitalizes on the developmental accomplishments within the domain of ANN to facilitate the instantiation of intricate SNN architectures. The translation of ANN architectures into SNN frameworks is one of the prevailing strategies for deriving operationally relevant deep-level SNN models. This method underscores the utilization of well-established network paradigms and theoretical constructs originating from the ANN domain, thereby expediting the realization of a biomimetic SNN model characterized by commendable efficacy and minimal power utilization.

The foundational underpinnings of the ANN-to-SNN algorithm emanate from the postulation that the dynamics inherent to a spiking neuron can be effectively emulated through the amalgamation of a linear neuron and a ReLU function. Additionally, to ensure a robust conversion from ANN to SNN, the architectural blueprint of the ANN necessitates adherence to the ensuing stipulations: 1) Enforcing strictly positive output values for all neurons. 2) Nullifying the influence of biases by setting them to a zero value. 3) Substituting the conventional max pooling with the utilization of average pooling.

In essence, the ANN-to-SNN algorithm circumvents the intricacies associated with training intricate deep SNN models. By harnessing the sophisticated gradient-based backpropagation mechanism inherent to the ANN framework, the intricate construction of a high-performing deep SNN model is actualized. Naturally, implementing such a conversion strategy entails a degree of performance compromise. Nevertheless, this performance deficit is progressively ameliorated through systematic refinement of the conversion algorithm.

#### 5) Surrogate Gradient Algorithm

In contrast to the ANN-to-SNN methodology, the surrogate gradient approach directly facilitates the training of intricate SNN architectures by iteratively refining synaptic weights through updates. However, the direct training of SNN models presents a formidable challenge owing to the inherent non-differentiability of spike sequences, a characteristic that impedes the conventional backpropagation process. Consequently, the fundamental premise of the surrogate gradient algorithm [70] revolves around the substitution of the spike function with a mathematically smooth and continuous surrogate function during the backpropagation phase, thereby effectuating the seamless transfer of gradients and optimization of synaptic weights. During the forward propagation phase, the spike function continues to be employed, enabling deep neuronal units to encode the spike sequences and thereby facilitate the acquisition of salient features. Within this paradigm, selecting diverse continuous surrogate functions is a pivotal consideration. Exemplars of such functions encompass the Sigmoid function (Sigmoid($x$)), the SoftSign function (SoftSign($x$)), and the Arctangent function (ATan($x$)), among others. Particularly, the ATan($x$) surrogate function holds widespread utilization, typified by its formulation as follows:

$$ATan(x) = \frac{1}{\pi} \arctan\left(\frac{\pi}{2}\alpha x\right) + \frac{1}{2} \quad (15)$$

Its gradient is:

$$\frac{d}{dx}ATan(x) = \frac{\alpha}{2\left(1 + \left(\frac{\pi}{2}\alpha x\right)^2\right)} \quad (16)$$

Through the application of surrogate gradient techniques, the intricate process of training deep SNN models becomes tractable, albeit with careful consideration of the surrogate function's characteristics and their influence on optimization dynamics.

IV. SNN-Based Deep Model for Intelligent IFD

This section delves into the research and utilization of SNN-based models within IFD. Primarily, we analyze the distinctive merits inherent to SNN, elucidating the rationale and imperative underpinning its application within IFD contexts. Subsequently, grounded in practical industrial scenarios, we expound upon the methodologies for deploying SNN-based models in IFD applications, thereby capitalizing upon their intrinsic advantages. Lastly, we explore extant IFD models, subjecting them to comprehensive scrutiny and synthesizing our analytical insights.

*A. Why use SNN for Fault Diagnosis*

As aforementioned, the SNN serves as an artificial intelligence paradigm emulating the operational dynamics of biological neurons. This emulation encompasses the intricate temporal characteristics of neurons, thereby enabling the assimilation of commendable attributes from the biological neural framework, encompassing energy efficiency, reasoning capacity, memory retention, and learning efficacy. Consequently, the SNN exhibits a discernible realm of promise for research and practical applications. Given the escalating prominence of limitations [71] within deep learning models about training expenses, generalization proficiencies, and interpretability, the realm of brain-inspired computation, akin to the SNN paradigm, has emerged as a burgeoning focal point for scholarly exploration. IFD applications based on deep models also face the above limitations, making it challenging to deploy deep models reliably in real industrial scenarios to achieve health monitoring of complex electromechanical equipment. In this milieu, the SNN model emerges as a viable prospect for ameliorating the shortcomings pervasive within deep models across multifarious dimensions. To this end, the SNN model is expected to improve the deficiencies of deep models in all aspects to promote the development of the field of fault diagnosis. In summary, SNN can shine in the field of diagnosis in the following aspects.

**1) Spatiotemporal Information Modeling.** Spatiotemporal information modeling plays a pivotal role in intelligent IFD, owing to the frequent manifestation of distinctive patterns and dynamic temporal changes within time series data in fault scenarios. SNNs, characterized by their inherent spiking encoding methodology, exhibit remarkable capabilities in encoding spatiotemporal information. By harnessing rate coding and temporal coding mechanisms, SNNs proficiently encapsulate intricate patterns and variations inherent in input data. Notably, research conducted by Fang et al. [72] underscores the SNN model's resemblance to a recurrent neural network architecture, wherein information updates involve the selective retention of present information while discarding a portion of past information at each time step. Consequently, the salient advantage of SNN-based spatiotemporal information modeling resides in its ability to precisely discern and forecast potential failure modes, effectively capturing subtle dynamic alterations within time series data.

**2) Sparsity and Energy Efficiency.** SNN simulates the pulse discharge process of neurons in biological neural networks, and the activities of neurons are usually sparse, which makes it have relatively low computational complexity and memory requirements and better energy efficiency. Such efficiency positions SNNs as promising candidates for applications in low-power devices and environments constrained by energy resources. In the domain of fault diagnosis, where terminal health monitoring platforms for electromechanical systems often rely on edge computing or embedded devices, the availability of limited computing resources and energy resources align seamlessly with the capabilities of the SNN model, instilling optimism regarding its potential to excel in such operational scenarios.

**3) Robustness and Noise Immunity.** The distinctive attributes encompassing sparsity and spatiotemporal modeling within the SNN framework endow it with formidable robustness and commendable immunity to the deleterious effects of noise and uncertainty present in input data. This resilience primarily emanates from the bedrock of SNN, namely, spiking encoding, which enables the model to withstand modest noise levels without compromising the fidelity of encoded feature information. Furthermore, the inherent property of sparsity empowers the SNN model to judiciously retain the most pertinent and consequential information while systematically expunging extraneous and superfluous data. This inherent trait amplifies the SNN model's potential in robust and precise health monitoring and fault diagnosis, especially within intricate and multifaceted operational settings.

**4) Event-Driven and Asynchronicity.** The distinctive hallmark of the SNN model resides in its event-driven and asynchronous characteristics. Within the framework of an event-driven model, the system eschews the conventional practice of awaiting periodically sampled continuous inputs. Instead, it promptly reacts to events as they manifest, demonstrating a heightened capacity to accommodate irregular time intervals and the asynchrony associated with signal occurrence. This kind of neural network performs better when dealing with discrete events and discontinuous signals, so it is more suitable for fault diagnosis of electromechanical equipment than the ANN model in some scenarios.

**5) Heterogeneous Data Processing.** In the context of fault diagnosis, incorporating diverse sensor data types, including but not limited to sound, vibration, and temperature, necessitates a robust approach to their processing. SNNs emerge as a versatile solution capable of flexibly managing this heterogeneity and seamlessly integrating disparate data streams within a unified spiking encoding framework, facilitating comprehensive analysis. This method uniformly encodes distinct sensor data types into neuromorphic representations, employing the spiking encoding framework to intricately merge and encode these data to extract and characterize the most valuable hybrid features.

In summary, the bionic characteristics and information encoding mechanism inherent in SNNs endow them with distinctive advantages that distinguish them from ANNs. Fusing these advantageous traits with real-world industrial contexts promises substantial application potential and holds the promise of catalyzing advancements within fault diagnosis.

*B. How to Apply SNN to Fault Diagnosis*

This section succinctly outlines the process of applying SNNs to develop an IFD model framework. Similar to the ANN model, SNN-based IFD models encompass the following fundamental steps: data preprocessing, model architecture design, model optimization, model evaluation and deployment. Below, we expound upon these four steps in meticulous detail.

**1) Data Processing.** Diverging from the ANN model paradigm, the SNN model emulates the spiking discharge and transmission mechanisms observed in biological neurons. Consequently, a critical preprocessing step involves the application of spiking encoding to transform continuous data into a spiking representation. The fundamental objective of spiking encoding lies in articulating data information through attributes such as pulse frequency and temporal dynamics. In doing so, SNNs can effectively discern and encapsulate essential temporal correlations and dynamics inherent in industrial signals, facilitating profound data encoding. The Poisson encoding method [73, 74] is the conventional approach for encoding continuous data into neuromorphic data, thereby creating spiking data representations. Nevertheless, contemporary research endeavors have explored innovative pathways wherein continuous data is directly utilized as input for the SNN model [75], often incorporating an internal spiking encoding layer to effectuate signal spiking encoding within the model.

**2) Model Design.** Developing an SNN model necessitates tailored design considerations contingent upon the research objectives and application context. The SNN model mainly has the following three model design directions: Bionic-based methods, deep model-based methods, and ANN-to-SNN-based methods. Bionic-based methods commit to crafting a model that adheres closely to the biological brain's neural network architecture. To formulate the model's structure in harmony with biological neural principles, it strives to simulate various facets of the biological neural network to the greatest extent possible. Deep model-based methods are dedicated to constructing an efficient and pragmatic SNN model, leveraging the architecture of a well-established ANN as a foundational template. By drawing upon the wealth of experience accrued in ANNs, this approach aims to devise an SNN model that exhibits superior performance. ANN-to-SNN-based methods seek to harness pre-existing research outcomes and exemplary network structures from the domain of ANNs to facilitate the creation of SNN models. This method directly converts accomplished ANN models into SNN models using conversion algorithms, thus inheriting the excellence of performance exhibited by the original ANN models. Though these three design schemes employ divergent implementation approaches, they share a common objective: developing an outstanding SNN model meticulously crafted to address practical problem-solving challenges.

**3) Model Optimization.** There are many optimization methods and training algorithms for SNN models. Appropriate optimization algorithms must be selected for different SNN models and scenarios. SNN models grounded in biomimetic principles predominantly lean towards optimization algorithms inspired by biological paradigms. For instance, Zhang et al. [76] proposed a biologically reasonable reward propagation algorithm, which can significantly reduce the computational cost of SNN. Conversely, SNN models rooted in deep learning principles typically gravitate towards optimization techniques founded on gradient backpropagation. For instance, Li et al. [77] proposed a new differentiable spike function based on finite difference gradient to train and optimize deep SNN models. The ANN-to-SNN approach initially employs gradient backpropagation to optimize the parameters of the ANN model. Subsequently, it necessitates additional algorithms to effectuate the conversion of ANN architectures into SNN structures. For instance, Chen et al. [78] proposed an adaptive threshold mechanism that improves the balance between SNN weights and thresholds, making the conversion from ANN to SNN better. Notably, owing to the distinctive characteristics of SNN models, existing optimization algorithms often exhibit varying degrees of limitations.

Consequently, developing innovative and effective optimization algorithms tailored for IFD applications holds the promise of harnessing the full potential of SNN models.

**4) Model Evaluation and Deployment.** The evaluation of SNN models closely mirrors the procedures adopted for ANN models. Evaluation results inform model refinement, enabling enhancements in accuracy and robustness through adjustments to network architecture, pulse encoding schemes, augmentation of training data, and related strategies. The deployment of SNN models encompasses two primary avenues: conventional hardware systems and neuromorphic hardware systems. While viable for SNN deployment, conventional hardware systems may only partially exploit the inherent advantages of SNN models, potentially limiting their potential gains in energy efficiency and other domains. In contrast, deploying SNN models in neuromorphic hardware systems offers the prospect of harnessing the model's capabilities more optimally. However, it is essential to acknowledge that the current landscape of neuromorphic hardware systems is still in its nascent research and development phase, posing challenges for large-scale deployment and utilization. Hence, selecting an appropriate deployment solution should be made judiciously, considering the specific characteristics and requirements of the given practical context.

Based on the above steps, developing a tailored SNN-based IFD model for real-world industrial applications is feasible. Nevertheless, it is imperative to acknowledge that this endeavor remains fraught with numerous challenges and limitations, which will be examined and discussed in Section V.

*C. SNN-Based Fault Diagnosis Model*

The remarkable attributes and distinctive advantages of SNNs have kindled growing interest among scholars in fault diagnosis, leading to a surge in related research endeavors and practical applications. Current scholarly efforts primarily revolve around developing efficacious SNN-based deep models and optimization algorithms tailored to address specific fault diagnosis challenges. For instance, Zuo et al. [79] pioneered the application of SNNs in bearing fault diagnosis, introducing a signal-to-spiking encoding method grounded in local mean decomposition. They augmented model training by employing an enhanced Tempotron learning rule. Subsequently, Zuo et al. [80] extended the SNN framework into a multi-layer deep model to achieve superior diagnostic performance. Liu et al. [81] contributed to the field by devising an event-driven spiking deep belief network model for robotic arm fault classification, featuring the reward-STDP learning rule for model training. Xu et al. [82] introduced a deep spiking residual shrinkage network, incorporating attention mechanisms and soft thresholding to heighten bearing fault recognition in noisy environments. Recent advancements by Wang et al. [83] ventured into the application of SNNs for health monitoring of autonomous vehicle sensors, introducing a membrane learnable mechanism within the SNN model, which exhibited robust performance in sensor health monitoring within open environments. Furthermore, Cao et al. [84] forged innovative ground by integrating SNNs with graph neural networks, yielding the Spike Graph Attention Network for intelligent IFD of planetary gearboxes, achieving synchronous encoding of spatiotemporal features of vibration signals. However, it is discernible that the exploration and application of SNNs within the fault diagnosis domain remain in nascent stages. Only a limited array of SNN-based diagnostic models have been devised to tackle specific engineering challenges. In the practical industrial sphere, a multitude of scenarios and tasks necessitate solutions, necessitating a more expansive and comprehensive exploration of SNNs to cater to diverse industrial needs.

V. CHALLENGES, SOLUTIONS AND OPPORTUNITIES

While the application of SNN models in fault diagnosis has yielded promising results, it still faces severe challenges and limitations that need to be solved and studied. In this section, we delve into the multifaceted challenges encountered by SNNs within the fault diagnosis domain and explore potential solutions and opportunities. This comprehensive survey provides valuable insights for future research in this critical field.

*A. Neuromorphic Datasets*

In practical industrial scenarios, data acquisition typically involves capturing diverse high-frequency or low-frequency sensor signals. Meanwhile, the SNN model is tailored to process spiking-based neuromorphic data. Hence, the imperative arises to develop robust algorithms facilitating the transformation of industrial data into a neuromorphic format. Many data transformation techniques have been proposed; however, it is essential to recognize that these existing methodologies primarily stem from image domains. In stark contrast, industrial sensor data presents complex temporal dynamic properties and faces the interference of various noises. For example, when a high-speed train is running, its speed, load, temperature, humidity, track status, etc., all change with time, and these changes in the external environment will also cause disturbances to the sensor monitoring data. Additionally, given the complex electromechanical nature of such devices, factors like friction and component collisions yield a cacophony of noise that adversely affects sensor signal monitoring. Consequently, developing innovative neuromorphic data transformation algorithms becomes imperative, necessitating carefully considering the unique characteristics inherent in industrial sensor data. Such algorithms should not only facilitate the encoding of intricate spatiotemporal dynamics within the signals but also possess the capability to effectively mitigate noise. Furthermore, it is noteworthy that within the realm of fault diagnosis, a dearth of neuromorphic industrial datasets exists. Consequently, establishing and disseminating open-source neuromorphic datasets tailored for fault diagnosis research is pivotal. The provision of such datasets holds the potential to significantly catalyze advancements in SNN-based fault diagnosis research.

*B. SNN-based Model Architecture*

The neural network architecture serves as the foundational framework and structural core of deep models, thereby positively influencing the model's performance. It is imperative to underscore that the seemingly boundless flexibility of model architecture presents a formidable challenge in the quest for optimal design. The extensive array of potential architectural configurations makes identifying an architecture ideally suited for fault diagnosis applications an intricate undertaking. As elucidated in Section IV, there are three directions for model architecture design: bionic-inspired methods, deep model-based methods, and ANN-to-SNN-based methods. However, in fault diagnosis applications, the uniqueness of data and the variability of application scenarios make model architecture design more complex. For example, the fault diagnosis field lacks the architectural design ideas of excellent ANN models and classic and practical ANN-based fault diagnosis models. This makes deep model-based methods and ANN-to-SNN-based model design schemes face significant challenges in fault diagnosis. Furthermore, the bionic-inspired method, characterized by its elevated complexity, moderate performance, and intricate optimization requirements, presents distinct challenges when applied to fault diagnosis scenarios. Therefore, in the future, it is a feasible solution to summarize the excellent design experience of ANN-based models and try to use these experiences in SNN-based models. At the same time, it is also necessary to consider the special properties of SNN models and fault diagnosis applications.

*C. Reliable Optimization Algorithms*

The SNN model uses pulse coding as its characteristic, which also leads to the problem of difficulty in optimization. Unlike ANN, information in SNN is represented by a sequence of impulses, which makes reliable gradient backpropagation very difficult. Consequently, optimizing SNN-based models has prompted the development of various algorithms. While these optimization techniques have shown promise in enhancing the performance of SNN models, they are not without limitations. For instance, the STDP algorithm struggles to optimize deep SNN models effectively. The surrogate gradient algorithm has made significant strides in incorporating gradient-based backpropagation within SNN optimization frameworks. However, its gradient information is inaccurate, which may lead to unpredictable deviations. Consequently, pursuing dependable optimization algorithms remains a primary objective within the SNN research community. Similarly, reliable optimization algorithms are also essential when applying SNN to fault diagnosis applications. The data dimension in the field of fault diagnosis is relatively low. Hence, the complexity requirements for the model are also relatively low, which makes the surrogate gradient algorithm perform well in fault diagnosis applications. The refinement of the surrogate gradient algorithm offers promising avenues for future research. Potential improvements include identifying more suitable gradient substitution functions and developing more efficient training strategies [85, 86] to make the SNN model more applicable to fault diagnosis application scenarios.

*D. Neuromorphic Hardware Systems*

In event-based SNN, the calculation is triggered by events; that is, the calculation is only performed when a spiking occurs. Given the inherently sparse nature of these spikes, SNNs exhibit exceptionally low computational energy consumption. This attribute renders SNNs particularly well-suited for deployment on energy-constrained platforms, including various micro-computing systems, IoT devices, and edge computing terminal devices. Within these systems, energy consumption serves as a paramount metric, and conventional ANN models are known to impose substantial resource demands. However, SNN models require neuromorphic hardware systems that match event-triggered computational models to fully exploit their advantages. Currently, building and utilizing specialized spike-based neuromorphic hardware systems is still in its early stages. Therefore, there are still many problems to be solved. Similarly, exploring effective and dependable neuromorphic hardware systems in fault diagnosis is paramount. The integration of such systems into diverse edge computing terminals holds the potential to significantly mitigate energy consumption, thereby enhancing device battery life and overall operational longevity.

*E. Model Interpretability*

Similar to ANN models, SNN models also grapple with limited interpretability. While SNNs faithfully replicate neuronal dynamics and exhibit substantial biological relevance, achieving interpretability, especially in deep SNN models, remains a formidable task when considered through a primarily biological lens. This challenge stems primarily from the incomplete elucidation of both the learning mechanisms and the intricacies of the biological brain. Moreover, profound disparities exist between deep-level SNNs and biological nervous systems, further complicating interpretability analyses. The proliferation of parameters in deep SNNs exacerbates the complexity of interpretability assessment. In applications such as fault diagnosis, model interpretability assumes paramount importance as stakeholders necessitate a comprehensive understanding of the model's learned representations and the rationale behind its decisions. To this end, exploring the interpretability of SNN models is also significant research. This pursuit may be approached from two aspects: 1) biologically based interpretable analysis. This method uses biological learning mechanisms as a template to build corresponding SNN models, using biological knowledge and theory to achieve considerable interpretability. 2) Emphasizing parameter optimization and feature learning analyses, delving into the optimization mechanisms governing SNN models, and dissecting the modes of feature acquisition, thereby enriching the interpretability of SNN models.

Additionally, leveraging Bayesian and similar methodologies can augment our comprehension of model uncertainty.

*F. Challenges and Difficulties of IFD*

In addition to the limitations of SNNs, the realm of IFD confronts a multitude of formidable challenges and intricacies. For example, 1) a considerable imbalance exists between fault and normal samples due to the sparsity of fault occurrence. 2) The rarity of industrial fault data and the high cost of data annotation require the model to achieve excellent performance on a small amount of annotated data sets. 3) Dynamic alterations in operating conditions or environmental factors introduce substantial perturbations to the underlying data feature distribution, posing a fundamental query: How to accurately diagnose the fault in unknown scenarios? 4) How to realize the identification and early warning of fault status in the scenario of only normal samples. Addressing these challenges mandates a deliberate and focused exploration of SNN model paradigms. This exploration may encompass the SNN model based on sample imbalance, the SNN model based on few-shot learning, the SNN model based on transfer learning, the SNN model based on unsupervised anomaly detection, etc. In summary, the SNN model has broad research prospects and practical value in fault diagnosis, which requires a lot of research and practice to promote the application and development of the SNN model in the industrial field.

## VI. CONCLUSIONS

SNN is a neural network system based on brain-like computing. Thanks to its unique advantages, it is known as the third-generation neural network. As the development of ANN-based deep learning models falls into a bottleneck, its limitations of resource dependence, data dependence, and limited generalization ability are always challenging to solve. This has led researchers to explore the research and application of SNN models in IFD to solve practical problems faced in industrial systems. However, there is currently a lack of comprehensive survey on SNN models in IFD to elaborate on their technical details, research status, limitations, and opportunities. In order to fill this gap, this paper systematically expounds the technical details of the SNN-based model, including the neuron model, information encoding mechanism, topology, and optimization algorithm, to answer what SNN is. Using SNN in the field of fault diagnosis and how to apply SNN, explaining the potential advantages of the SNN model in fault diagnosis applications. Subsequently, this article analyzes the existing SNN-based fault diagnosis models and systematically answers the challenges, solutions, and opportunities of SNN in IFD. In summary, this review systematically introduces the research progress, limitations, and prospects of SNN models in fault diagnosis, which is expected to provide valuable guidance for research in this field.

> REPLACE THIS LINE WITH YOUR MANUSCRIPT ID NUMBER (DOUBLE-CLICK HERE TO EDIT) <

129, pp. 4757-4769, 2022.
[24] Y. Wei and H. Wang, "Wavelet integrated attention network with multi-resolution frequency learning for mixed-type wafer defect recognition," *Eng. Appl. Artif. Intel.*, vol. 121, pp. 105975, 2023.
[25] H. Wang, Z. Liu, D. Peng and M.J. Zuo, "Interpretable convolutional neural network with multilayer wavelet for Noise-Robust Machinery fault diagnosis," *Mech. Syst. Signal Pr.*, vol. 195, pp. 110314, 2023.
[26] H. Meng, M. Geng and T. Han, "Long short-term memory network with Bayesian optimization for health prognostics of lithium-ion batteries based on partial incremental capacity analysis," *Reliab. Eng. Syst. Safe.*, vol. 236, pp. 109288, 2023.
[27] Z. Yang, B. Xu, W. Luo and F. Chen, "Autoencoder-based representation learning and its application in intelligent fault diagnosis: A review," *Measurement*, vol. 189, pp. 110460, 2022.
[28] J. Jiao, M. Zhao, J. Lin and K. Liang, "A comprehensive review on convolutional neural network in machine fault diagnosis," *Neurocomputing*, vol. 417, pp. 36-63, 2020.
[29] A. Krizhevsky, I. Sutskever and G.E. Hinton, "ImageNet classification with deep convolutional neural networks," *Commun. ACM*, vol. 60, no. 6, pp. 84-90, 2017.
[30] K. He, X. Zhang, S. Ren and J. Sun, "Deep residual learning for image recognition," in *Proc. IEEE Conference on Computer Vision and Pattern Recognition (CVPR)*, 2016, pp. 770-778.
[31] J. Zhu, Q. Jiang, Y. Shen and C. Qian, et al., "Application of recurrent neural network to mechanical fault diagnosis: a review," *Journal of Mechanical Science and Technology*, vol. 2, no. 36, pp. 527-542, 2022.
[32] J. Shi, D. Peng, Z. Peng and Z. Zhang, et al., "Planetary gearbox fault diagnosis using bidirectional-convolutional LSTM networks," *Mech. Syst. Signal Pr.*, vol. 162, pp. 107996, 2022.
[33] H. Wang, Z. Liu and T. Ai, "Long-range dependencies learning based on non-local 1D-convolutional neural network for rolling bearing fault diagnosis.," *Journal of Dynamics, Monitoring and Diagnostics*, vol. 1, pp. 148-159, 2022.
[34] H. Wang and Y.F. Li, "Iterative Error Self-Correction for Robust Fault Diagnosis of Mechanical Equipment With Noisy Label," *IEEE T. Instrum. Meas.*, vol. 71, pp. 1-13, 2022.
[35] W. Luo and H. Wang, "Composite Wafer Defect Recognition Framework based on Multi-View Dynamic Feature Enhancement with Class-Specific Classifier," *IEEE T. Instrum. Meas.*, pp. 1, 2023.
[36] A. Tavanaei, M. Ghodrati, S.R. Kheradpisheh and T. Masquelier, et al., "Deep learning in spiking neural networks," *Neural Networks*, vol. 111, pp. 47-63, 2019.
[37] A. Taherkhani, A. Belatreche, Y. Li and G. Cosma, et al., "A review of learning in biologically plausible spiking neural networks," *Neural Networks*, vol. 122, pp. 253-272, 2020.
[38] S. Zheng, L. Qian, P. Li and C. He, et al., "An Introductory Review of Spiking Neural Network and Artificial Neural Network: From Biological Intelligence to Artificial Intelligence," [Online]. Available: https://arxiv.org/abs/2204.07519.
[39] D. Zhang, T. Zhang, S. Jia and Q. Wang, et al., "Recent Advances and New Frontiers in Spiking Neural Networks," [Online]. Available: https://arxiv.org/abs/2204.07050.
[40] J. Pei, L. Deng, S. Song and M. Zhao, et al., "Towards artificial general intelligence with hybrid Tianjic chip architecture," *Nature*, vol. 572, no. 7767, pp. 106, 2019.
[41] K. Roy, A. Jaiswal and P. Panda, "Towards spike-based machine intelligence with neuromorphic computing," *Nature*, vol. 575, no. 7784, pp. 607-617, 2019.
[42] B. Yin, F. Corradi and S.M. Bohté, "Accurate and efficient time-domain classification with adaptive spiking recurrent neural networks," *Nature Machine Intelligence*, vol. 3, no. 10, pp. 905-913, 2021.
[43] W. Fang, Z. Yu, Y. Chen and T. Huang, et al., "Deep residual learning in spiking neural networks," *Advances in Neural Information Processing Systems*, vol. 34, pp. 21056-21069, 2021.
[44] E. Zio, "Prognostics and Health Management (PHM): Where are we and where do we (need to) go in theory and practice," *Reliab. Eng. Syst. Safe.*, vol. 218, pp. 108119, 2022.
[45] X. Dai and Z. Gao, "From Model, Signal to Knowledge: A Data-Driven Perspective of Fault Detection and Diagnosis," *IEEE T. Ind. Inform.*, vol. 9, no. 4, pp. 2226-2238, 2013.
[46] Z. Zhu, Y. Lei, G. Qi and Y. Chai, et al., "A review of the application of deep learning in intelligent fault diagnosis of rotating machinery," *Measurement*, vol. 206, pp. 112346, 2023.
[47] Z. He, H. Shao, P. Wang and J.J. Lin, et al., "Deep transfer multi-wavelet auto-encoder for intelligent fault diagnosis of gearbox with few target training samples," *Knowl.-Based Syst.*, vol. 191, pp. 105313, 2020.
[48] H. Wang, Z. Liu, D. Peng and Z. Cheng, "Attention-guided joint learning CNN with noise robustness for bearing fault diagnosis and vibration signal denoising," *ISA T.*, vol. 128, pp. 470-484, 2022.
[49] C. Li, S. Li, H. Wang and F. Gu, et al., "Attention-based deep meta-transfer learning for few-shot fine-grained fault diagnosis," *Knowl.-Based Syst.*, vol. 264, pp. 110345, 2023.
[50] J. Yao and T. Han, "Data-driven lithium-ion batteries capacity estimation based on deep transfer learning using partial segment of charging/discharging data," *Energy*, vol. 271, pp. 127033, 2023.
[51] X. Yang, Z. Song, I. King and Z. Xu, "A Survey on Deep Semi-Supervised Learning," *IEEE T. Knowl. Data En.*, vol. 35, no. 9, pp. 8934-8954, 2023.
[52] G. He, Y. Pan, X. Xia and J. He, et al., "A Fast Semi-Supervised Clustering Framework for Large-Scale Time Series Data," *IEEE Transactions on Systems, Man, and Cybernetics: Systems*, vol. 51, no. 7, pp. 4201-4216, 2021.
[53] X. Tao, X. Gong, X. Zhang and S. Yan, et al., "Deep Learning for Unsupervised Anomaly Localization in Industrial Images: A Survey," *IEEE T. Instrum. Meas.*, vol. 71, pp. 1-21, 2022.
[54] X. Liu, F. Zhang, Z. Hou and L. Mian, et al., "Self-Supervised Learning: Generative or Contrastive," *IEEE T. Knowl. Data En.*, vol. 35, no. 1, pp. 857-876, 2023.
[55] A.L. Hodgkin and A.F. Huxley, "A quantitative description of membrane current and its application to conduction and excitation in nerve," *The Journal of physiology*, vol. 117, no. 4, pp. 500, 1952.
[56] W. Gerstner and W.M. Kistler, "Spiking neuron models: Single neurons, populations, plasticity,", 2002.
[57] E.M. Izhikevich, "Simple model of spiking neurons," *IEEE Transactions on Neural Networks*, vol. 14, no. 6, pp. 1569-1572, 2003.
[58] P.A. Cariani, "Temporal codes and computations for sensory representation and scene analysis," *IEEE Transactions on Neural Networks*, vol. 15, no. 5, pp. 1100-1111, 2004.
[59] Q. Yu, H. Tang, K.C. Tan and H. Li, "Rapid Feedforward Computation by Temporal Encoding and Learning With Spiking Neurons," *IEEE T. Neur. Net. Lear.*, vol. 24, no. 10, pp. 1539-1552, 2013.
[60] I. Sporea and A. Grüning, "Supervised Learning in Multilayer Spiking Neural Networks," *Neural Comput.*, vol. 25, no. 2, pp. 473-509, 2013.
[61] A. Shrestha, K. Ahmed, Y. Wang and D.P. Widemann, et al., "A spike-based long short-term memory on a neurosynaptic processor," in *Proc. IEEE/ACM International Conference on Computer-Aided Design (ICCAD)*, 2017, pp. 631-637.
[62] S.R. Kheradpisheh, M. Ganjtabesh, S.J. Thorpe and T. Masquelier, "STDP-based spiking deep convolutional neural networks for object recognition," *Neural Networks*, vol. 99, pp. 56-67, 2018.
[63] A. Tavanaei and A.S. Maida, "Multi-layer unsupervised learning in a spiking convolutional neural network," in *Proc. International Joint Conference on Neural Networks (IJCNN)*, 2017, pp. 2023-2030.
[64] N. Caporale and Y. Dan, "Spike timing‐dependent plasticity: a Hebbian learning rule," *Annu. Rev. Neurosci.*, vol. 31, pp. 25-46, 2008.
[65] Y. Dan and M. Poo, "Spike timing-dependent plasticity: from synapse to perception," *Physiol. Rev.*, vol. 86, no. 3, pp. 1033-1048, 2006.
[66] F. Ponulak and A. Kasiński, "Supervised learning in spiking neural networks with ReSuMe: sequence learning, classification, and spike shifting," *Neural Comput.*, vol. 22, no. 2, pp. 467-510, 2010.
[67] S.M. Bohte, J.N. Kok and H. La Poutré, "Error-backpropagation in temporally encoded networks of spiking neurons," *Neurocomputing*, vol. 48, no. 1, pp. 17-37, 2002.
[68] Y. Cao, Y. Chen and D. Khosla, "Spiking deep convolutional neural networks for energy-efficient object recognition," *Int. J. Comput. Vision*, vol. 113, pp. 54-66, 2015.
[69] P.U. Diehl, D. Neil, J. Binas and M. Cook, et al., "Fast-classifying, high-accuracy spiking deep networks through weight and threshold balancing," in *Proc. International Joint Conference on Neural Networks (IJCNN)*, 2015, pp. 1-8.
[70] E.O. Neftci, H. Mostafa and F. Zenke, "Surrogate Gradient Learning in Spiking Neural Networks: Bringing the Power of Gradient-Based Optimization to Spiking Neural Networks," *IEEE Signal Proc. Mag.*, vol. 36, no. 6, pp. 51-63, 2019.
[71] M.M. Waldrop, "What are the limits of deep learning?" *Proceedings of the National Academy of Sciences*, vol. 116, no. 4, pp. 1074-1077, 2019.
[72] W. Fang, Z. Yu, Y. Chen and T. Masquelier, et al., "Incorporating learnable membrane time constant to enhance learning of spiking neural